\documentclass[lettersize,journal]{IEEEtran}
\usepackage{amsmath,amsfonts}
\usepackage{algorithmic}
\usepackage{algorithm}
\usepackage{array}
\usepackage[caption=false,font=normalsize,labelfont=sf,textfont=sf]{subfig}
\usepackage{textcomp}
\usepackage{stfloats}
\usepackage{url}
\usepackage{verbatim}
\usepackage{graphicx}
\usepackage{cite}
\usepackage{color}
\usepackage{epstopdf}
\DeclareMathOperator{\tr}{tr}

\newtheorem{lemma}{Lemma}
\newtheorem{theorem}{Theorem}

\begin{document}

\title{Clustering Result Re-guided Incomplete Multi-view Spectral Clustering}

\author{Jun~Yin,~Runcheng~Cai, and~Shiliang~Sun,~\IEEEmembership{Senior Member,~IEEE}

\thanks{Jun Yin and Runcheng Cai are with the College of Information Engineering, Shanghai Maritime University, Shanghai 201306, China (e-mail: junyin@shmtu.edu.cn; 202030310093@stu.shmtu.edu.cn).}
\thanks{Shiliang Sun is with the Department of Automation, Shanghai Jiao Tong University, Shanghai 200240, China (e-mail: shiliangsun@gmail.com).}
\thanks{\emph{Corresponding author: Shiliang Sun}}

}



\maketitle

\begin{abstract}
Incomplete multi-view spectral clustering generalizes spectral clustering to multi-view data and simultaneously realizes the partition of multi-view data with missing views. For this category of method, $K$-means algorithm needs to be performed to generate the clustering result after the procedure of feature extraction.  More importantly, the connectivity of samples reflected by the clustering result is not utilized effectively. To overcome these defects, we propose Clustering Result re-Guided Incomplete Multi-view Spectral Clustering (CRG\_IMSC). CRG\_IMSC obtains the clustering result directly by imposing nonnegative constraint to the extracted feature. Furthermore, it constructs the connectivity matrix according to the result of spectral clustering, and minimizes the residual of self-representation based on the connectivity matrix. A novel iterative algorithm using multiplicative update is developed to solve the optimization problem of CRG\_IMSC, and its convergence is proved rigorously. On benchmark datasets, for multi-view data, CRG\_IMSC performs better than state-of-the-art clustering methods, and the experimental results also demonstrate the convergence of CRG\_IMSC algorithm.
\end{abstract}


\section{Introduction}
\IEEEPARstart{C}{lustering} is a classical unsupervised machine learning method, which partitions data without label. It is widely applied in text analysis \cite{ref1}, image segmentation \cite{ref2} and autonomous driving \cite{Maria2021}, etc. In the past, many excellent clustering methods were developed, such as $K$-means \cite{ref3,ref4}, hierarchical clustering \cite{ref5,ref6} and spectral clustering \cite{ref7,ref8}.

Nowadays, information is often collected from multiple sources and represented by multiple views. For example, a news article is reported by different media, and a video is composed of picture and audio. To partition multi-view data, Multi-View Clustering (MVC) methods are developed \cite{ref9}. MVC can be generally divided into three categories, i.e., Matrix Factorization-based MVC (MFMVC), Graph-based MVC (GMVC) and Deep Learning-based MVC (DLMVC).

MFMVC methods derive the low dimensional representation of each view via matrix factorization. Liu et al. \cite{ref10} utilized Nonnegative Matrix Factorization (NMF) to generate a common representation and designed an effective normalization approach to make different views have similar clustering results. Cai et al. \cite{ref11} learned the low dimensional representation of multi-view data by constrained NMF, and clustered samples with the same label into the clustering prototypes. Chang et al. \cite{ref12} derived the latent features by concept factorization and combined manifold learning to keep geometric structure of the data.

GMVC methods obtain potential relationships between samples in each view by constructing similarity graphs. Hu et al. \cite{ref13} combined the objectives of NMF and spectral clustering, and constructed a parameter-free model. Chen et al. \cite{ref14} inferred connections among all instances by learning latent embedding space and designed the similarity graph with the potential embedding to derive the cluster indicator matrix. Liang et al. \cite{ref15} defined a min-max framework for robust graph learning and transformed it to a differentiable and convex objective to solve the non-convex problem.

DLMVC methods extract meaningful information from original data by deep neural network. Li et al. \cite{ref16} extracted latent features by deep auto-encoders and used adversarial strategy to obtain data distribution and derive the latent representation. Trosten et al. \cite{ref17} combined representations of multiple views with a linear combinations learned by a view-prioritization mechanism and introduced a contrastive alignment module to preserve the local geometric structure and view invariance. Xu et al. \cite{ref18} derived the embedded features of multiple views by deep auto-encoders and learned the feature representations and cluster assignments by collaborative training scheme.

However, in the real world, some data from partial views is possibly lost due to human or machine errors. To partition multi-view data with missing views, the incomplete multi-view clustering methods are proposed. In general, these methods inherit the advantage of MVC methods with the enhancement of handling missing data. There are two common types of incomplete multi-view clustering approaches. One type is to recover missing data. Yang et al. \cite{ref19} recovered missing data by fitting the potential structure of each view and simultaneously utilized the complementary information and view-specific information by matching graphs from different views. Yin and Sun \cite{ref20} recovered missing views to exploit the relation of samples sufficiently. The recovered views and latent representation interact in iterations to obtain more accurate clustering results. Liu et al. \cite{ref21} imputed kernel matrix with missing elements in the procedure of pursuing clustering results, and their method can run even though all the kernel matrices are incomplete. Wen et al. \cite{ref22} inferred missing data by locality preserving reconstruction and employed the reverse graph regularization to learn a unified graph to keep the common local structure of different views. Xue et al. \cite{ref23} used auto-encoder to recover missing information, and then extracted data structure by adaptive graph learning and graph convolution. Xu et al. \cite{xu2021} recovered missing views with the latent representation of multi-view data, and evaluated the reconstruction by the generative adversarial network.

Another type is to analyze the incomplete multi-view data directly and consider the negative effect of missing data. Li et al. \cite{ref24} sought a latent space where different views of the same sample are close and simultaneously the data in single view have perfect clustering structure. Hu and Chen \cite{DAIMC} employed the instance alignment to learn the latent features, and aligned the basis matrices of multiple views to reduce the damage of missing views. Zhuge \cite{ref25} measured the relationships of samples by similarity matrix, and derived the latent representation and the common probability label matrix simultaneously. Liang et al. \cite{ref27} learned the sample-level auto-weight for graph fusion, and considered different contributions of multiple views and the impact of missing data. Wen et al. \cite{ref28} employed view-specific encoder to get high-level information and utilized a fusion layer to get the common representation of multi-view data with missing views. Besides, a clustering layer is adopted to derive the optimal clustering result. Chen et al. \cite{Chen2023} proposed a Low-Rank Tensor Learning for incomplete multi-view clustering, which explores the global and local structures of the data simultaneously and utilizes the high-order correlations of different views. In addition, to handle the problems of new samples and unbalanced information of different views, Deng et al. \cite{Deng2023} developed a graph regularized Projective consensus representation learning model.

As a graph based clustering method, spectral clustering \cite{ref29} is successfully applied to multi-view clustering and incomplete multi-view clustering, and obtains excellent performance \cite{ref34,ref39}. In multi-view spectral clustering, for each view, the similarity graph is constructed to preserve the relationship of samples in the procedure of seeking the common representation of multi-view data. However, the similarity graph can only represent the pairwise similarity of samples. The relationship between sample and class is not explicitly considered, which is extremely important to clustering. Additionally, after seeking feature representation, $K$-means algorithm is usually performed to realize the partition of the new representation, which artificially separates feature representation and clustering, and simultaneously increases the complexity of clustering method.

In this paper, we propose Clustering Result re-Guided Incomplete Multi-view Spectral Clustering (CRG\_IMSC). CRG\_IMSC obtains the clustering result from the representation directly by imposing nonnegative constraint on the representation, avoiding $K$-means after feature extraction. The nonnegative representation reflects the relationship of sample and class, which is adopted to construct the connectivity matrix of samples. The connectivity matrix is then employed in self-representation, and the minimization of the residual of self-representation re-guides the incomplete multi-view spectral clustering. The objective function of CRG\_IMSC is solved by an iterative algorithm, and a novel multiplicative update rule is proposed to complete alternative iteration. The convergence of the iterative algorithm is proved in theory and displayed in the experiments. CRG\_IMSC is compared with state-of-the-art clustering methods on public multi-view datasets. Experimental results show the effectiveness of CRG\_IMSC.

\section{Related Works}
\subsection{Multi-view spectral clustering}
Spectral clustering is derived from graph theory, which can keep the topological structure of the data. Multi-view spectral clustering extends spectral clustering to the multi-view domain, exploring relationships between samples by preserving graph structures in each view. Zhu et al. \cite{ref30} learned the common similarity matrix for clustering, and removed the redundancy and the noise of the original data. Sharma and Seal et al. \cite{ref31} constructed a self-adaptive mixture similarity with kernel distance and Jeffrey-divergence, and incorporated it into multi-view spectral clustering model. Huang et al. \cite{ref32} applied deep neural networks to multi-view spectral clustering, and integrated the local invariant information and the consistency of different views into one model. Li et al. \cite{ref33} incorporated heterogeneous features with manifold structure and used bipartite graph to simulate the similarity graph to increase computational efficiency. To handle the out-of-sample extension problem in multi-view spectral clustering, Shi et al. \cite{Shi2023} designed a self-adaption strategy to construct graph and fuse the graphs of different views.

Co-regularized multi-view spectral clustering \cite{ref34} is a classic multi-view spectral clustering method which utilizes the co-regularized scheme to reduce the divergence between different views to obtain a common representation. Its objective function is defined as
\begin{equation}\begin{split}
&\underset{U^{v},U}{\text{max}}\sum_{v=1}^{V}\left(\begin{split} &\tr(U^{v}S_{nor}^{v}U^{v^{T}})\\ &+\lambda ^v\tr(U^{v^{T}}U^{v}U^{T}U) \end{split} \right)
\\&s.t. \quad U^{v}U^{v^{T}}=I, UU^{T}=I,
\end{split}\end{equation}
where $U^{v}\in \mathbb{R}^{k\times n}$ is the feature representation of the $v$th view, $U$ is the common representation, $n$ and $k$ are the number of samples and the number of clusters, respectively, $\lambda^v$ is the weight coefficient of the $v$th view,
$S_{nor}^{v}=(D^{v})^{-1/2}S^{v}(D^{v})^{-1/2}$ denotes the normalized graph Laplacian of the $v$th view, $S^{v}\in \mathbb{R}^{n\times n}$ is the similarity matrix of the $v$th view, and $D^{v}$ is a diagonal matrix with the diagonal element $ D_{ii}^{v} = {\textstyle \sum_{j=1}^{n}} S_{ij}^{v}$.

\subsection{Incomplete multi-view spectral clustering}
Incomplete multi-view spectral clustering approach has also received attention from researchers when faced with incomplete data. Yin and Sun \cite{ref35} calculated the cosine similarity for the purpose of maintaining the local structure of the original data, and integrated the local structure preserving and matrix factorization into a unified model. Wang et al. \cite{ref37} transformed the missing data problem to missing similarity problem, and reduced the perturbation risk across different views to get a common representation. Wen et al. \cite{ref38} obtained the local information from the incomplete multi-view data, and combined view-specific information and consistent information of multiple views. Yin et al. \cite{ref36} connected all instances and reconstructed missing data by anchor points, and then performed spectral clustering with the anchor graph instead of the traditional similarity graph to reduce time complexity. Xia et al. \cite{xia2022} employed tensor technique to impute the similarity matrix of incomplete data, making that the similarity matrix can maintain the relationship of the existing data.

Adaptive Graph Completion based Incomplete Multi-view Clustering (AGC\_IMC) \cite{ref39} is a representative incomplete multi-view spectral clustering method, which integrates within-view preservation, between-view inferring and common representation learning into a framework, and introduces adaptive weights for different views. Its objective function is formulated as
\begin{equation}\begin{split}
&\underset{S^{v},U,B,\alpha ^{v}}{\text{min}} \sum \limits _{v = 1}^V(\alpha ^{v})^{r} \left( \begin{split}&\lambda _{1}  \left\Vert{S^{v}} - \sum \limits _{i = 1,i \ne v}^V {{S^i}{B_{i,v}}}\right\Vert_{F}^{2}
\\ &+\left\Vert(S^{v}-\tilde{S}^{v})\odot {W^{v}}\right\Vert_{F}^{2} \\&+\lambda _{2}\tr({U{L_{{S^{v}}}}{U^T}}) \end{split}\right)
 \\& s.t. \quad 0 \le {S^{v}} \le 1,{S^v}^T\emph{1} = \emph{1}, S_{i,i}^{v} = 0, U{U^T} = I,
 \\ &0 \le {B_{i,v}} \le 1, \sum \limits_{i=1,i\ne v}^V {{B_{i,v}}} = 1, {B_{v,v}} = 0,
 \\&  \sum \limits_{v = 1}^V {{\alpha ^{v}}} = 1, 0 \le {\alpha ^{v}} \le 1,
\end{split}\end{equation}
where $U$ is the unified representation, $S^v$ is the completed similarity matrix, $\tilde{S}^{v}$ is the extended similarity matrix whose elements corresponding to missing samples are set as 0 and other elements are the similarities of existing samples, $L_{S^v}=D^{v}-(S^v+{S^v}^T)/2$ is the Laplacian matrix, $D^{v}$ is a diagonal matrix with the diagonal element $D_{ii}^{v} = {\textstyle \sum_{j=1}^{n}} (S_{ij}^{v}+S_{ji}^{v})/2$, $B\in\mathbb{R}^{V\times V}$ is a self-representation matrix, $\alpha ^{v}$ is the weight of the $v$th view, $r$ is a smooth parameter to balance the weight distribution, $\odot$ is the Hadamard product, and $W^{v}$ indicates the sample existence condition, $W_{i,j}^{v}=1$ if the $v$th view of the $i$th sample and the $j$th sample both exist, otherwise $W_{i,j}^{v}=0$. For the extended similarity matrix $\tilde{S}^v$, its elements corresponding to the missing data are set as 0.

\section{Proposed Method}
In this section, we first formulate the objective of CRG\_IMSC. Then we give optimization of the objective function and analyze convergence of the optimization algorithm.
\subsection{Objective Function}
Traditionally, after obtaining the representation of multi-view data, multi-view spectral clustering performs $K$-means to obtain the clustering result of the representation. It increases the computational complexity and separates representation leaning and clustering, which may makes the obtained representation not appropriate for clustering. To solve this problem, we integrate clustering procedure into the objective function of multi-view spectral clustering, and obtain the clustering result from this objective directly. We define the objective as

\begin{equation}\begin{split}
\min_{U^{v},U}&\sum_{v=1}^V\left(\begin{split}&\tr\left(U^{v}L^{v}{U^{v}}^T\right)\\
&+\beta\left\|{M^{v}}^TU^TUM^{v}-{U^{v}}^T{U^{v}}\right\|_F^2\end{split}\right)\\
s.t. \quad & U^{v}{U^{v}}^T=I^{v}, U\geq0,
\end{split}\label{ref3}\end{equation}
where $\beta$ is a  hyper-parameter, $L^{v}=(D^{v})^{-1/2}(D^{v}-S^{v})(D^{v})^{-1/2}$  is the normalized Laplacian matrix of the $v$th view, $S^v$ is the similarity matrix, $D^{v}$ is a diagonal matrix with the diagonal element $D_{ii}^{v} = {\textstyle \sum_{j=1}^{n}}(S_{ij}^{v}+S_{ji}^{v})/2$, $I^{v}\in\mathbb{R}^{n^v\times n^v}$ is the identity matrix, $U^{v}\in\mathbb{R}^{k\times n^v}$ is the representation of the $v$th view, $U\in\mathbb{R}^{k\times n}$ is the common representation of all the views, $n^v$ and $n$ are the number of existing samples of the $v$th view and the number of all the samples respectively, and $M^{v}\in\mathbb{R}^{n\times n^v}$ is the mark matrix of the existing samples of the $v$th view, which is obtained by deleting the columns of the identity matrix $I\in\mathbb{R}^{n\times n}$ corresponding to the missing samples of the $v$th view. It is easy to know that $UM^{v}$ is the common representation of $X^{v}$, and the minimization of $\left\|{M^{v}}^TU^TUM^{v}-{U^{v}}^T{U^{v}}\right\|_F^2$ guarantees the consistency between $U$ and each $U^{v}$. In Eq.~\eqref{ref3}, we impose nonnegative constraint on the representation $U$. Then, the $l$th row and $i$th column element of $U$, i.e., $U_{li}$, can be approximately regarded as the probability of the $i$th sample $x_i$ belonging to the $l$th cluster, and $x_i$ is assigned to the cluster $c$ with the maximum probability, i.e., $c=\underset{l}{\arg\max}U_{li}$. In fact, the common representation $U$ is just the clustering result.

The clustering result $U$ reflects the connection of cluster and sample. For instance, $U_{li}$ is the connection of the $l$th cluster and the $i$th sample. A column of $U$ represents the connections between one sample and all the clusters. Suppose that $U_i$ and $U_j$ are the $i$th and the $j$th column of $U$ respectively. Then the connection of the $i$th sample and the $j$th sample can be calculated by $U_i^TU_j$, and $U^TU$ is exactly the connectivity matrix of all the samples. The connectivity matrix has strong relationship with the principle of self-representation. In self-representation, we hope that a sample can be more represented by other samples which have strong connection with it. This objective can be formulated as
\begin{equation}\begin{split}
\min_U& \sum_{v=1}^V\left(\left\|X^{v}-X^{v}\left(\left({M^{v}}^TU^TUM^{v}\right)\odot A^{v}\right)\right\|_F^2\right),
\end{split}\label{ref4}\end{equation}
where $A^{v}\in\mathbb{R}^{n^v\times n^v}$ is a matrix whose diagonal elements are $0$ and other elements are $1$. With $A^{v}$, we make a sample represented by other samples except itself.

\begin{figure*}[!h]
\centering
\includegraphics[width=18cm]{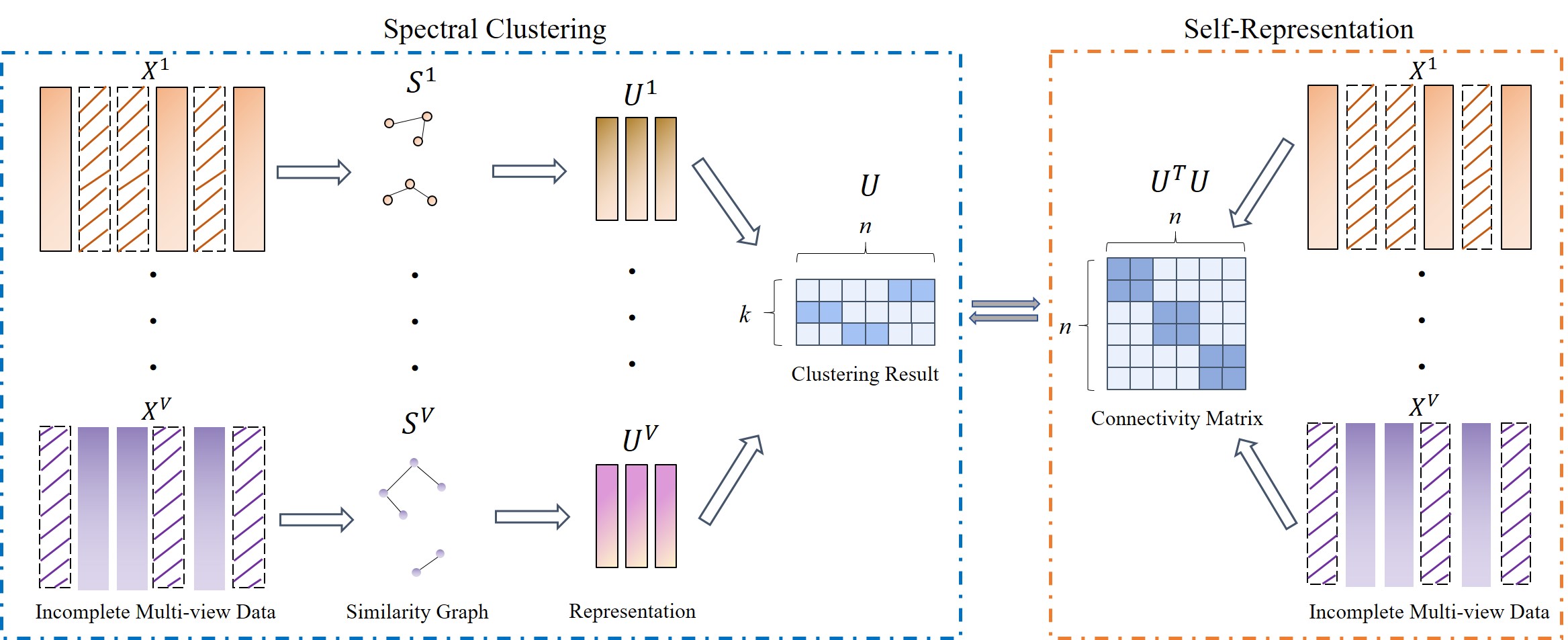}
\caption{The framework of CRG\_IMSC: In the left side, spectral clustering is performed on the incomplete multi-view data to obtain the clustering result $U$. Then the connectivity matrix which reflects the relationship of different samples can be calculated by $U^TU$. In the right side, based on the connectivity matrix, self-representation is performed to re-guide the incomplete multi-view spectral clustering.}
\label{fig1}
\end{figure*}

Combining the objectives \eqref{ref3} and \eqref{ref4}, we have the final objective of CRG\_IMSC as
\begin{equation}\begin{split}
\min_{U^{v},U}&\sum_{v=1}^V\left(\begin{split}
&\left\|X^{v}-X^{v}\left(\left({M^{v}}^TU^TUM^{v}\right)\odot A^{v}\right)\right\|_F^2\\
&+\alpha\tr\left(U^{v}L^{v}{U^{v}}^T\right)\\
&+\beta\left\|{M^{v}}^TU^TUM^{v}-{U^{v}}^T{U^{v}}\right\|_F^2\end{split}\right)\\
s.t. \quad & U^{v}{U^{v}}^T=I^{v}, U\geq0,
\end{split}\label{ref5}\end{equation}
where $\alpha$ is a hyper-parameter. In Eq.~\eqref{ref5}, the clustering result $U$ obtained by incomplete multi-view spectral clustering is also influenced by self-representation. By minimizing the residual of self-representation, the clustering result can re-guide incomplete multi-view spectral clustering. The framework of CRG\_IMSC is displayed in Fig.~\ref{fig1}.

\subsection{Optimization}
Considering that there are two variables in the objective of CRG\_IMSC, we solve it by alternating optimization strategy. The update procedure is presented as follows.

\textbf{Step 1:} Fix $U$ and update $U^{v}$.\\

Since $U$ is fixed, we need to solve the optimization problem as
\begin{equation}\begin{split}
\min_{U^{v}}&\sum_{v=1}^V\left(\begin{split}&\alpha\tr\left(U^{v}L^{v}{U^{v}}^T\right)\\
&+\beta\left\|{M^{v}}^TU^TUM^{v}-{U^{v}}^T{U^{v}}\right\|_F^2\end{split}\right)\\
s.t. \quad & U^{v}{U^{v}}^T=I^{v}.
\end{split}\label{ref6}\end{equation}
The optimization problem \eqref{ref6} can be transformed to
\begin{equation}\begin{split}
\min_{U^{v}}&\sum_{v=1}^V\tr\left(U^{v}\left(\alpha L^{v}-2\beta{M^{v}}^TU^TUM^{v}\right){U^{v}}^T\right)\\
s.t. \quad & U^{v}{U^{v}}^T=I^{v}.
\end{split}\label{ref7}\end{equation}
This optimization problem can be solved by the eigen-decomposition of $\alpha L^{v}-2\beta{M^{v}}^TU^TUM^{v}$. The rows of $U^{v}$ is composed by the eigen-vectors corresponding to the $k$ smallest eigenvalues.

\textbf{Step 2:} Fix $U^{v}$ and update $U$.\\

With fixed $U^{v}$, we can transformed the optimization problem \eqref{ref5} to
\begin{equation}\begin{split}
\min_{U}&L(U)=\sum_{v=1}^V\left(\begin{split}
&\tr\left(\left(Q^{v}\odot A^{v}\right)^TP^{v}\left(Q^{v}\odot A^{v}\right)\right)\\
&-2\tr\left(P^{v}\left(Q^{v}\odot A^{v}\right)\right)\\
&+\beta\tr\left({Q^{v}}^TQ^{v}\right)-2\beta\tr\left({Q^{v}}^T{U^{v}}^TU^{v}\right)
\end{split}\right)\\
s.t. &\quad U\geq0,
\end{split}\label{ref8}\end{equation}
where $P^{v}={X^{v}}^TX^{v}$ and $Q^{v}={M^{v}}^TU^TUM^{v}$. We can solve this minimization problem by the multiplicative update algorithm, which is obtained by auxiliary function method (see the section of convergence analysis). $U$ is updated as
\begin{equation}U_{li}\xrightarrow{}U_{li}\sqrt[4] {\frac{\sum\limits_{v=1}^V\left(\begin{split}&2U{M^{v}}\left(P^{v}\odot A^{v}\right){M^{v}}^T\\
&+2\beta UM^{v}\left({U^{v}}^TU^{v}\right)^+{M^{v}}^T\end{split}\right)_{li}}
{\sum\limits_{v=1}^V\left(\begin{split}&UM^{v}\left(P^{v}\left(Q^{v}\odot A^{v}\right)\odot A^{v}\right){M^{v}}^T\\
&+UM^{v}\left(\left(Q^{v}\odot A^{v}\right)P^{v}\odot A^{v}\right){M^{v}}^T\\
&+2\beta UM^{v}Q^{v}{M^{v}}^T\\
&+2\beta UM^{v}\left({U^{v}}^TU^{v}\right)^-{M^{v}}^T\end{split}\right)_{li}}},\label{ref10}\end{equation}
where
\begin{equation}
\left({U^{v}}^TU^{v}\right)^+=\frac{\left|{U^{v}}^TU^{v}\right|+{U^{v}}^TU^{v}}{2},
\end{equation}
\begin{equation}
\left({U^{v}}^TU^{v}\right)^-=\frac{\left|{U^{v}}^TU^{v}\right|-{U^{v}}^TU^{v}}{2}
\end{equation}
and
\begin{equation}
{U^{v}}^TU^{v}=\left({U^{v}}^TU^{v}\right)^+-\left({U^{v}}^TU^{v}\right)^-.
\end{equation}
Here we suppose that the multi-view data $X^{v}$ is nonnegative. For negative $X^{v}$, we can transform it to be nonnegative by normalization. With the iterative steps above, the CRG\_IMSC algorithm is listed in Algorithm \ref{algo1}.

\renewcommand{\algorithmicrequire}{\textbf{Input:}}  
\renewcommand{\algorithmicensure}{\textbf{Output:}} 
\begin{algorithm}[!h]
\caption{CRG\_IMSC algorithm}
\begin{algorithmic}[1] 
\REQUIRE ~~\\ 
Incomplete multi-view data $X^{v}|_{v=1}^V$; hyper-parameters $\alpha$ and $\beta$.
\ENSURE ~~\\ 
The clustering result $U$.
\STATE Initialize $U$.
\REPEAT
\STATE Fix $U$, solve the eigen-decomposition problem in Eq.~\eqref{ref7}, update $U^v$ with the eigenvectors corresponding to $k$ smallest eigenvalues.
\STATE Fix $U^{v}|_{v=1}^V$ , update $U$ by Eq.~\eqref{ref10}.
\UNTIL convergence.
\end{algorithmic}
\label{algo1}
\end{algorithm}

In the beginning of the algorithm, for simplicity, the random matrix can be used to initialize $U$. However, the random initialization will lead to the fluctuation of clustering results. In order to obtain relatively stable result, in this paper, NMF is employed to initialize $U$. Suppose $Z^v\in\mathbb{R}^{k\times n^v}$ is the coefficient matrix obtained by performing NMF in $X^v$. Then $U$ can be initialized as $\frac{1}{V}\sum_{v=1}^VZ^v{M^v}^T+0.1E$, where $E\in\mathbb{R}^{k\times n}$ is a matrix with all elements equal to 1.

\subsection{Convergence Analysis}

\begin{lemma}
 \begin{equation}\begin{split}
&G(U,\hat{U})\\
=&\sum\limits_{v=1}^V\left(\begin{split}
&\sum_{li}\frac{\left(\hat{U}M^{v}\left(\left(\hat{Q}^{v}\odot A^{v}\right)P^{v}\odot A^{v}\right){M^{v}}^T\right)_{li}U_{li}^4}{2\hat{U}_{li}^3}\\
&+\sum_{li}\frac{\left(\hat{U}M^{v}\left(P^{v}\left(\hat{Q}^{v}\odot A^{v}\right)\odot A^{v}\right){M^{v}}^T\right)_{li}U_{li}^4}{2\hat{U}_{li}^3}\\
&-2\sum_{lij}\left(\begin{split}
&\left(M^{v}\left(P^{v}\odot A^{v}\right){M^{v}}^{T}\right)_{ij}\\
&\times\hat{U}_{li}\hat{U}_{lj}\left(1+\ln{\frac{U_{li}U_{lj}}{\hat{U}_{li}\hat{U}_{lj}}}\right)\end{split}\right)\\
&+\beta\sum_{li}\frac{\left(\hat{U}M^{v}\hat{Q}^{v}{M^{v}}^T\right)_{li}U_{li}^4}{\hat{U}_{li}^3}\\
&+2\beta\sum_{li}\frac{\left(\hat{U}M^{v}\left({U^{v}}^TU^{v}\right)^-{M^{v}}^{T}\right)_{li}\left({U}_{li}^4+\hat{U}_{li}^4\right)}{2\hat{U}_{li}^3}\\
&-2\beta\sum_{lij}\left(\begin{split}&\left(M^{v}\left({U^{v}}^TU^{v}\right)^+{M^{v}}^{T}\right)_{ij}\\
&\times\hat{U}_{li}\hat{U}_{lj}\left(1+\ln{\frac{U_{li}U_{lj}}{\hat{U}_{li}\hat{U}_{lj}}}\right)\end{split}\right)
\end{split}
\right)
 \end{split}\end{equation}
 is an auxiliary function of the objective $L(U)$ in Eq.~\eqref{ref8},
 where $\hat{Q}^{v}={M^{v}}^T\hat{U}^T\hat{U}M^{v}$.
\label{lem1}\end{lemma}

To prove the convergence of CRG\_IMSC algorithm, we first demonstrate that the objective of CRG\_IMSC is monotonically decreasing with the update of $U$ in Eq.~\eqref{ref10}. We employ the auxiliary function method \cite{Daniel2001}. If a function $G(Z,\hat{Z})$ satisfies $G(Z,\hat{Z})\geq L(Z)$ and $G(Z,Z)=L(Z)$ for any $Z$ and $\hat{Z}$, it is an auxiliary function of $L(Z)$ . According to the above definition, we have Lemma \ref{lem1}.

\begin{IEEEproof}
Since $Q^{v}$ and $A^{v}$ are symmetric matrices, the function $L(U)$ in Eq.~\eqref{ref8} can be transformed to
\begin{equation}\begin{split}
&L(U)\\
=&\sum_{v=1}^V\left(\begin{split}
&\frac{\tr\left(\left(\left(Q^{v}\odot A^{v}\right)P^{v}\odot A^{v}\right)Q^{v}\right)}{2}\\
&+\frac{\tr\left(Q^{v}\left(P^{v}\left(Q^{v}\odot A^{v}\right)\odot A^{v}\right)\right)}{2}\\
&-2\tr\left(\left(P^{v}\odot A^{v}\right)Q^{v}\right)\\
&+\beta\tr\left({Q^{v}}Q^{v}\right)+2\beta\tr\left({Q^{v}}\left({U^{v}}^TU^{v}\right)^-\right)\\
&-2\beta\tr\left({Q^{v}}\left({U^{v}}^TU^{v}\right)^+\right)
\end{split}\right)\\
=&\sum_{v=1}^V\left(\begin{split}
&\frac{\tr\left(UM^{v}\left(\left(Q^{v}\odot A^{v}\right)P^{v}\odot A^{v}\right){M^{v}}^TU^T\right)}{2}\\
&+\frac{\tr\left(UM^{v}\left(P^{v}\left(Q^{v}\odot A^{v}\right)\odot A^{v}\right){M^{v}}^TU^T\right)}{2}\\
&-2\tr\left(UM^{v}\left(P^{v}\odot A^{v}\right){M^{v}}^TU^T\right)\\
&+\beta\tr\left(UM^{v}{M^{v}}^TU^TUM^{v}{M^{v}}^TU^T\right)\\
&+2\beta\tr\left(UM^{v}\left({U^{v}}^TU^{v}\right)^-{M^{v}}^TU^T\right)\\
&-2\beta\tr\left(UM^{v}\left({U^{v}}^TU^{v}\right)^+{M^{v}}^TU^T\right)
\end{split}\right)
\end{split}\label{ref15}\end{equation}
As $M^{v}\left(P^{v}+I\right){M^{v}}^T$ is symmetric matrix, using Lemma \ref{lem2} (see below) and setting  $A\xrightarrow{}M^{v}\left(P^{v}+I\right){M^{v}}^T$, $B\rightarrow{U}$, we have
\begin{equation}\begin{split}
&\tr\left(UM^{v}\left(P^{v}+I\right)Q^{v}\left(P^{v}+I\right){M^{v}}^TU^T\right)\\
=&\tr\left(UM^{v}\left(P^{v}+I\right){M^{v}}^TU^TUM^{v}\left(P^{v}+I\right){M^{v}}^TU^T\right)\\
\leq&\sum_{li}\frac{\left(\hat{U}M^{v}\left(P^{v}+I\right){M^{v}}^T\hat{U}^T\hat{U}M^{v}\left(P^{v}+I\right){M^{v}}^T\right)_{li}U_{li}^4}{\hat{U}_{li}^3}\\
=&\sum_{li}\frac{\left(\hat{U}M^{v}\left(P^{v}+I\right)Q^v\left(P^{v}+I\right){M^{v}}^T\right)_{li}U_{li}^4}{\hat{U}_{li}^3}.
\end{split}\label{ref16}\end{equation}
Then we can easily obtain
\begin{equation}\begin{split}
&\frac{\tr\left(UM^{v}\left(\left(Q^{v}\odot A^{v}\right)P^{v}\odot A^{v}\right){M^{v}}^TU^T\right)}{2}\\
+&\frac{\tr\left(UM^{v}\left(P^{v}\left(Q^{v}\odot A^{v}\right)\odot A^{v}\right){M^{v}}^TU^T\right)}{2}\\
\leq&\sum_{li}\frac{\left(\hat{U}M^{v}\left(\left(\hat{Q}^{v}\odot A^{v}\right)P^{v}\odot A^{v}\right){M^{v}}^T\right)_{li}U_{li}^4}{2\hat{U}_{li}^3}\\
+&\sum_{li}\frac{\left(\hat{U}M^{v}\left(P^{v}\left(\hat{Q}^{v}\odot A^{v}\right)\odot A^{v}\right){M^{v}}^T\right)_{li}U_{li}^4}{2\hat{U}_{li}^3}.
\end{split}\label{ref17}\end{equation}
Similarly, using Lemma \ref{lem2} and setting $A\rightarrow{}M^{v}{M^{v}}^T$, we have
\begin{equation}\begin{split}
&\tr\left(UM^{v}{M^{v}}^TU^TUM^{v}{M^{v}}^TU^T\right)\\
\leq&\sum_{li}\frac{\left(\hat{U}M^{v}{M^{v}}^T\hat{U}^T\hat{U}M^{v}{M^{v}}^T\right)_{li}U_{li}^4}{\hat{U}_{li}^3}\\
=&\sum_{li}\frac{\left(\hat{U}M^{v}\hat{Q}^{v}{M^{v}}^T\right)_{li}U_{li}^4}{\hat{U}_{li}^3}.
\end{split}\label{ref18}\end{equation}
Since $M^{v}\left({U^{v}}^TU^{v}\right)^-{M^{v}}^T$ is symmetric matrix, according to Lemma \ref{lem3} (see below) and with $A\xrightarrow{}M^{v}\left({U^{v}}^TU^{v}\right)^-{M^{v}}^T$, $B\rightarrow{U}$, there holds
\begin{equation}\begin{split}
&\tr\left(UM^{v}\left({U^{v}}^TU^{v}\right)^-{M^{v}}^TU^T\right)\\
\leq&\sum_{li}\frac{\left(\hat{U}M^{v}\left({U^{v}}^TU^{v}\right)^-{M^{v}}^{T}\right)_{li}\left({U}_{li}^4+\hat{U}_{li}^4\right)}{2\hat{U}_{li}^3}.
\end{split}\label{ref19}\end{equation}
Using the inequality $z\geq1+\ln{z}$, which holds for any $z>0$, we obtain
\begin{equation}\begin{split}
&\tr\left(UM^{v}\left(P^{v}\odot A^{v}\right){M^{v}}^TU^T\right)\\
=&\sum_{lij}\left(M^{v}\left(P^{v}\odot A^{v}\right){M^{v}}^{T}\right)_{ij}{U}_{li}{U}_{lj}\\
\geq&\sum_{lij}\left(M^{v}\left(P^{v}\odot A^{v}\right){M^{v}}^{T}\right)_{ij}
\hat{U}_{li}\hat{U}_{lj}\left(1+\ln{\frac{U_{li}U_{lj}}{\hat{U}_{li}\hat{U}_{lj}}}\right).
\end{split}\label{ref20}\end{equation}
and
\begin{equation}\begin{split}
&\tr\left(UM^{v}\left({U^{v}}^TU^{v}\right)^+{M^{v}}^TU^T\right)\\
=&\sum_{lij}\left(M^{v}\left({U^{v}}^TU^{v}\right)^+{M^{v}}^{T}\right)_{ij}{U}_{li}{U}_{lj}\\
\geq&\sum_{lij}\left(M^{v}\left({U^{v}}^TU^{v}\right)^+{M^{v}}^{T}\right)_{ij}\hat{U}_{li}\hat{U}_{lj}\left(1+\ln{\frac{U_{li}U_{lj}}{\hat{U}_{li}\hat{U}_{lj}}}\right)
\end{split}\label{ref21}\end{equation}
Combining Eq.~\eqref{ref17}, Eq.~\eqref{ref18}, Eq.~\eqref{ref19}, Eq.~\eqref{ref20} and Eq.~\eqref{ref21}, we have $G(U,\hat{U})\geq L(U)$. $G(U,U)=L(U)$ is obviously holds. Therefore, $G(U,\hat{U})$ is an auxiliary function of $L(U)$.
\end{IEEEproof}

\begin{lemma} For any matrices $A\geq0$, $B\geq0$ and $\hat{B}\geq0$, if $A$ is symmetric, there holds
\begin{equation}
\tr\left(BAB^TBAB^T\right)\leq\sum_{li}\frac{\left(\hat{B}A\hat{B}^T\hat{B}A\right)_{li}B_{li}^4}{\hat{B}_{li}^3}.
\label{ref22}\end{equation}
\label{lem2}\end{lemma}
\begin{IEEEproof}
Let $B_{li}=c_{li}\hat{B_{li}}$. The left-hand side and the right-hand side of the inequality \eqref{ref22} can be explicitly written as
\begin{equation}\begin{split}
\tr\left(BAB^TBAB^T\right)
=\sum_{lijoqr}\hat{B}_{lj}A_{jo}\hat{B}_{qo}\hat{B}_{qr}A_{ri}\hat{B}_{li}c_{lj}c_{qo}c_{qr}c_{li}
\end{split}\label{ref23}\end{equation}
and
\begin{equation}
\sum_{li}\frac{\left(\hat{B}A\hat{B}^T\hat{B}A\right)_{li}B_{li}^4}{\hat{B}_{li}^3}
=\sum_{lijoqr}\hat{B}_{lj}A_{jo}\hat{B}_{qo}\hat{B}_{qr}A_{ri}\hat{B}_{li}c_{li}^4.
\label{ref24}\end{equation}
Because $A$ is symmetric, there holds
\begin{equation}\begin{split}
&\sum_{lijoqr}\hat{B}_{lj}A_{jo}\hat{B}_{qo}\hat{B}_{qr}A_{ri}\hat{B}_{li}c_{li}^4\\
=&\sum_{lijoqr}\frac{\hat{B}_{lj}A_{jo}\hat{B}_{qo}\hat{B}_{qr}A_{ri}\hat{B}_{li}\left(c_{li}^4+c_{qr}^4\right)}{2}\\
\geq&\sum_{lijoqr}\hat{B}_{lj}A_{jo}\hat{B}_{qo}\hat{B}_{qr}A_{ri}\hat{B}_{li}\left(c_{li}^2c_{qr}^2\right)\\
=&\sum_{lijoqr}\frac{\hat{B}_{lj}A_{jo}\hat{B}_{qo}\hat{B}_{qr}A_{ri}\hat{B}_{li}\left(c_{li}^2c_{qr}^2+c_{lj}^2c_{qo}^2\right)}{2}\\
\geq&\sum_{lijoqr}\hat{B}_{lj}A_{jo}\hat{B}_{qo}\hat{B}_{qr}A_{ri}\hat{B}_{li}c_{lj}c_{qo}c_{qr}c_{li}.
\end{split}\label{ref25}\end{equation}
With Eq.~\eqref{ref23}, Eq.~\eqref{ref24} and inequality \eqref{ref25}, we can obtain inequality \eqref{ref22}.
\end{IEEEproof}

\begin{lemma} For any matrices $A\geq0$, $B\geq0$ and $\hat{B}\geq0$, if $A$ is symmetric, there holds
\begin{equation}
\tr\left(BAB^T\right)\leq\sum_{li}\frac{\left(\hat{B}A\right)_{li}\left(B_{li}^4+\hat{B}_{li}^4\right)}{2\hat{B}_{li}^3}.
\label{ref26}\end{equation}
\label{lem3}\end{lemma}
\begin{IEEEproof}
Let $B_{li}=c_{li}\hat{B_{li}}$. We can write the left-hand side and the right-hand side of Eq.~\eqref{ref26} as
\begin{equation}
\tr\left(BAB^T\right)=\sum_{lij}\hat{B}_{lj}A_{ji}\hat{B}_{li}c_{li}c_{lj}
\label{ref27}\end{equation}
and
\begin{equation}
\sum_{li}\frac{\left(\hat{B}A\right)_{li}\left(B_{li}^4+\hat{B}_{li}^4\right)}{2\hat{B}_{li}^3}=\sum_{lij}\frac{\hat{B}_{lj}A_{ji}\hat{B}_{li}\left(c_{li}^4+1\right)}{2}.
\label{ref28}\end{equation}
With the matrix $A$ symmetric, we have
\begin{equation}\begin{split}
&\sum_{lij}\frac{\hat{B}_{lj}A_{ji}\hat{B}_{li}\left(c_{li}^4+1\right)}{2}\\
=&\sum_{lij}\hat{B}_{lj}A_{ji}\hat{B}_{li}\left(\frac{c_{li}^4+c_{lj}^4}{4}+\frac{1}{2}\right)\\
\geq&\sum_{lij}\frac{\hat{B}_{lj}A_{ji}\hat{B}_{li}\left(c_{li}^2c_{lj}^2+1\right)}{2}\geq\sum_{lij}\hat{B}_{lj}A_{ji}\hat{B}_{li}c_{li}c_{lj}.
\end{split}\label{ref29}\end{equation}
Using Eq.~\eqref{ref27}, Eq.~\eqref{ref28} and inequality \eqref{ref29}, we can obtain inequality \eqref{ref26}.
\end{IEEEproof}

With Lemma \ref{lem1}, we give the following theorem, which demonstrates the objective of CRG\_IMSC is decreasing with the update of $U$ in Eq.~\eqref{ref10}.
\begin{theorem}
The objective of CRG\_IMSC $L(U)$ is decreasing if $U$ is updated by Eq.~\eqref{ref10}.
\label{the1}\end{theorem}
\begin{IEEEproof}
The derivative of $G(U,\hat{U})$ with respect to $U_{li}$ is
\begin{equation}\begin{split}
&\frac{\partial{G(U,\hat{U})}}{\partial{U_{li}}}\\
=&\sum\limits_{v=1}^V\left(\begin{split}
&\frac{2\left(\hat{U}M^{v}\left(\left(\hat{Q}^{v}\odot A^{v}\right)P^{v}\odot A^{v}\right){M^{v}}^T\right)_{li}U_{li}^3}{\hat{U}_{li}^3}\\
&+\frac{2\left(\hat{U}M^{v}\left(P^{v}\left(\hat{Q}^{v}\odot A^{v}\right)\odot A^{v}\right){M^{v}}^T\right)_{li}U_{li}^3}{\hat{U}_{li}^3}\\
&-\frac{4\left(\hat{U}M^{v}\left(P^{v}\odot A^{v}\right){M^{v}}^{T}\right)_{li}\hat{U}_{li}}{{U}_{li}}\\
&+\frac{4\beta\left(\hat{U}M^{v}\hat{Q}^{v}{M^{v}}^T\right)_{li}U_{li}^3}{\hat{U}_{li}^3}\\
&+\frac{4\beta\left(\hat{U}M^{v}\left({U^{v}}^TU^{v}\right)^-{M^{v}}^{T}\right)_{li}{U}_{li}^3}{\hat{U}_{li}^3}\\
&-\frac{4\beta\left(\hat{U}M^{v}\left({U^{v}}^TU^{v}\right)^+{M^{v}}^{T}\right)_{li}\hat{U}_{li}}{U_{li}}
\end{split}\right).
\end{split}\end{equation}
It is easy to verify that the Hessian matrix of $G(U,\hat{U})$ containing second derivatives is a diagonal matrix with positive elements. Thus $G(U,\hat{U})$ is a convex function of $U$. To find the minimum of $G(U,\hat{U})$, we set $\frac{\partial{G(U,\hat{U})}}{\partial{U_{li}}}=0$, and then Eq.~\eqref{ref10} is obtained. Suppose that $U^{(t+1)}$ and $U^{(t)}$ is $U$ of the $(t)$th iteration and the $(t+1)$th iteration respectively, and $U^{(t+1)}=\min\limits_{U}G(U,U^{(t)})$. According to the definition of auxiliary function, we have $L(U^{(t+1)})\leq G(U^{(t+1)},U^{(t)})\leq G(U^{(t)},U^{(t)})=L(U^{(t)})$.
\end{IEEEproof}

It is easy to see that the objective function of CRG\_IMSC is non-negative, which means that it has a lower bound. In the iteration algorithm of CRG\_IMSC, the update of $U^v$ obviously decreases the objective of CRG\_IMSC. According to Algorithm \ref{the1}, the update of $U$ also decreases the objective. The objective of CRG\_IMSC is monotonically decreasing in the update procedure. Based on the above analysis, we can conclude that CRG\_IMSC algorithm will converge.

\subsection{Complexity Analysis}
The time complexity of CRG\_IMSC algorithm is determined by iterative update of $U$ and $U^v$. In step 3 of Algorithm \ref{algo1}, for the update of $U$, the most time-consuming operations are matrix multiplication and eigen-decomposition, and the time complexities of them are $\mathcal{O}(\sum_{v=1}^Vnn^vn^v)$ and $\mathcal{O}(\sum_{v=1}^V(n^v)^3)$, respectively. In step 4 of Algorithm \ref{algo1}, for the update of $U^v$, the most time-consuming operation is matrix multiplication, and its time complexity is $\mathcal{O}(\sum_{v=1}^V((n^v)^3+knn^v))$. Considering $n>n^v>k$, we can obtain that the total time complexity of CRG\_IMSC algorithm is $\mathcal{O}(T\sum_{v=1}^Vnn^vn^v)$, where $T$ is the iterative times.

\subsection{Relation to other multi-view clustering methods}
For the objective of CRG\_IMSC in Eq.~\eqref{ref5}, the first term is self-representation used to make a sample more represented by other samples from the same cluster. If the first term is removed, this method turns into a common multi-view spectral clustering method. If we let the hyper-parameters $\alpha=0$ and $\beta=0$, only the self-representation constraint is reserved. In fact, now the method becomes a multi-view clustering method based on a special NMF, where the data matrix $X^v$ is decomposed and the basis matrix is $X^v$ itself. Different from previous multi-view clustering methods based on NMF, CRG\_IMSC can obtain the clustering results from the coefficient matrix directly, without performing K-means after NMF.

\section{Experiments}
In this section, we evaluate the effectiveness of CRG\_IMSC on four public multi-view datasets. Four widely used clustering validity indexes, i.e., Normalized Mutual Information (NMI), Accuracy, F-score, and Precision are employed to measure the clustering performance.
\subsection{Datasets}
\begin{table}[!h]
\renewcommand{\arraystretch}{1.3}
\centering
\caption{Statistic of Datasets}
\centering
\begin{tabular}{ccccc}
\hline
\bfseries Dataset & Size & Class & View & Dimension  \\
\hline
Wikipedia & 2866 & 10 & 2 & 10/128 \\
3Sources & 416 & 6 & 3 & 3068/3631/3560 \\
BBC & 2225 & 5 & 4 & 4659/4633/4665/4684 \\
BBCSport & 737 & 5 & 4 & 1991/2063/2113/2158 \\
\hline
\end{tabular}
\label{tab:data}
\end{table}
Four benchmark datasets are summarized in table ~\ref{tab:data}. In the following, we describe them in detail.\\
\textbf{Wikipedia}\cite{Costa2014}: Wikipedia consists of 2866 documents of 10 classes. These documents are collected from ``Wikipedia featured articles''. Each document contains text and image, which are two views of the document. \\
\textbf{3Sources}\cite{greene2009}: 3sources consists of 416 news stories of 6 classes collected from 3 media, i.e., BBC, Reuters, and The Guardian, which form 3 views of one story. Not all the 416 stories are reported by 3 media, the numbers of stories from 3 media are 294, 302 and 352, respectively. \\
\textbf{BBC}\cite{greene2006}: BBC consists of 2225 documents of 5 classes collected from the BBC news. 4 views of BBC are generated by segmenting the document. The number of documents in 4 views are 1543, 1524, 1574 and 1549, respectively. \\
\textbf{BBCSport}\cite{greene2006}: BBCSport consists of 737 documents of 5 classes collected from the BBC Sport. 4 views of BBCSport are generated by segmenting the documnet. The number of documents in views are 397, 410, 437 and 432, respectively.

\begin{table*}[!h]
\renewcommand{\arraystretch}{1.3}
\centering
\caption{NMI(\%)(mean and standard deviation) under different missing rates on Wikipedia dataset}
\footnotesize
\begin{tabular}{cccccc}
\hline
\bfseries Method\verb|\|MR & 10\% & 30\% & 50\% & 70\% & 90\% \\
\hline
BSV & \textbf{52.23}$\pm$0.29 & \textbf{44.55}$\pm$0.40 & \textbf{39.05}$\pm$0.38 & \textbf{33.52}$\pm$0.73 & 27.38$\pm$0.65 \\
MultiNMF & 36.44$\pm$0.95 & 32.80$\pm$0.44 & 28.12$\pm$1.27 & 24.66$\pm$1.28 & 20.41$\pm$1.19 \\
Spec-Pair & 39.31$\pm$1.47 & 35.48$\pm$0.81 & 30.79$\pm$2.18 & 24.18$\pm$2.41 & 16.91$\pm$5.24 \\
Spec-Cent & 44.36$\pm$0.40 & 38.21$\pm$0.45 & 34.02$\pm$0.76 & 30.34$\pm$0.89 & 26.36$\pm$0.90 \\
PVC & 50.69$\pm$2.90 & 40.89$\pm$6.23 & 34.11$\pm$2.92 & 27.73$\pm$2.44 & 22.76$\pm$2.13 \\
DAIMC & 45.83$\pm$0.81 & 31.48$\pm$0.80 & 15.77$\pm$1.59 & 5.50$\pm$1.06 & 2.49$\pm$0.51 \\
SRLC & 37.44$\pm$1.78 & 32.59$\pm$1.87 & 28.34$\pm$1.54 & 24.68$\pm$1.46 & 21.24$\pm$1.37 \\
UEAF & 48.24$\pm$0.41 & 41.96$\pm$0.49 & 35.80$\pm$0.48 & 30.90$\pm$0.88 & 26.11$\pm$0.83\\
AGC\_IMC & 47.99$\pm$1.04 & 42.08$\pm$1.04 & 35.72$\pm$0.78 & 31.17$\pm$1.40 & \textbf{27.99}$\pm$1.09\\
SAGF\_IMC & 31.49$\pm$4.57 & 30.56$\pm$2.59 & 25.25$\pm$1.97 & 21.66$\pm$2.53 & 18.21$\pm$0.96\\
LRTL & 34.02$\pm$1.62 & 28.83$\pm$1.63 & 23.50$\pm$2.15 & 18.63$\pm$1.02 &  17.83$\pm$1.20 \\
PIMVC & 45.43$\pm$0.69 & 40.18$\pm$0.94 & 35.41$\pm$0.95 & 29.63$\pm$1.42 & 21.11$\pm$0.75\\
CRG\_IMSC & 49.22$\pm$0.59 & 43.71$\pm$1.31 & 38.83$\pm$0.62 & 33.30$\pm$0.76 & 26.79$\pm$0.68 \\
\hline
\end{tabular}
\label{tab:wiki1}
\end{table*}

\begin{table*}[!h]
\renewcommand{\arraystretch}{1.3}
\centering
\caption{ACC(\%)(mean and standard deviation) under different missing rates on Wikipedia dataset}
\footnotesize
\begin{tabular}{cccccc}
\hline
\bfseries Method\verb|\|MR & 10\% & 30\% & 50\% & 70\% & 90\% \\
\hline
BSV & 56.13$\pm$0.33 & 49.47$\pm$0.81 & 45.43$\pm$0.65 & 39.72$\pm$1.00 & 33.43$\pm$0.73 \\
MultiNMF & 46.81$\pm$1.60 & 45.10$\pm$0.99 & 40.77$\pm$1.76 & 38.19$\pm$1.28 & 34.33$\pm$1.77 \\
Spec-Pair & 49.30$\pm$1.70 & 46.55$\pm$1.30 & 42.74$\pm$2.19 & 35.38$\pm$2.50 & 27.92$\pm$4.59 \\
Spec-Cent & 50.67$\pm$0.48 & 47.00$\pm$0.66 & 43.90$\pm$0.61 & 40.53$\pm$1.26 & 36.01$\pm$0.78 \\
PVC & 55.93$\pm$3.63 & 46.92$\pm$5.41 & 41.55$\pm$2.08 & 36.84$\pm$2.01 & 33.21$\pm$1.46 \\
DAIMC & 54.35$\pm$1.12 & 44.70$\pm$0.81 & 30.55$\pm$1.86 & 19.27$\pm$1.36 & 16.72$\pm$0.95 \\
SRLC & 48.44$\pm$1.06 & 45.75$\pm$0.88 & 42.39$\pm$0.84 & 39.36$\pm$1.18 & 36.14$\pm$0.77 \\
UEAF & 52.58$\pm$0.42 & 49.76$\pm$0.52 & 45.94$\pm$0.59 & 42.23$\pm$0.68 & 37.58$\pm$0.94\\
AGC\_IMC & 50.50$\pm$1.72 & 45.34$\pm$1.80 & 40.54$\pm$1.24 & 37.27$\pm$1.80 & 32.82$\pm$0.87\\
SAGF\_IMC & 35.52$\pm$3.65 & 36.21$\pm$2.39 & 33.72$\pm$2.47 & 30.86$\pm$2.53 & 29.65$\pm$1.46\\
LRTL & 45.52$\pm$1.43 & 41.40$\pm$1.41 & 37.74$\pm$1.23 & 33.51$\pm$0.75 & 32.97$\pm$0.87\\
PIMVC & 52.50$\pm$0.43 & 49.74$\pm$0.78 & 46.98$\pm$0.74 & 43.05$\pm$1.48 & 37.29$\pm$0.81\\
CRG\_IMSC & \textbf{57.27}$\pm$1.21 & \textbf{53.50}$\pm$1.87 & \textbf{50.72}$\pm$0.63 & \textbf{44.27}$\pm$1.04 & \textbf{38.51}$\pm$0.34\\
\hline
\end{tabular}
\label{tab:wiki2}
\end{table*}

\begin{table*}[!h]
\renewcommand{\arraystretch}{1.3}
\centering
\caption{F-score(\%)(mean and standard deviation) under different missing rates on Wikipedia dataset}
\footnotesize
\begin{tabular}{cccccc}
\hline
\bfseries Method\verb|\|MR & 10\% & 30\% & 50\% & 70\% & 90\% \\
\hline
BSV & 47.79$\pm$0.31 & 37.49$\pm$0.74 & 30.82$\pm$0.31 & 26.02$\pm$0.32 & 22.59$\pm$0.14 \\
MultiNMF & 35.16$\pm$1.02 & 31.69$\pm$0.41 & 27.72$\pm$0.75 & 25.04$\pm$0.75 & 21.95$\pm$1.01 \\
Spec-Pair & 38.97$\pm$1.42 & 34.85$\pm$0.92 & 31.03$\pm$1.47 & 25.43$\pm$1.96 & 19.99$\pm$2.96 \\
Spec-Pair & 38.97$\pm$1.42 & 34.85$\pm$0.92 & 31.03$\pm$1.47 & 25.43$\pm$1.96 & 19.99$\pm$2.96 \\
Spec-Cent & 41.97$\pm$0.43 & 36.63$\pm$0.62 & 32.00$\pm$0.61 & 27.95$\pm$0.72 & 23.69$\pm$0.50 \\
PVC & \textbf{49.14}$\pm$3.15 & 38.34$\pm$4.71 & 31.65$\pm$1.28 & 26.38$\pm$1.85 & 22.59$\pm$0.77 \\
DAIMC & 45.43$\pm$0.94 & 32.48$\pm$0.70 & 20.30$\pm$1.10 & 15.58$\pm$0.43 & 15.79$\pm$0.17 \\
SRLC & 37.31$\pm$1.34 & 33.23$\pm$0.80 & 29.29$\pm$0.81 & 25.76$\pm$1.08 & 22.88$\pm$0.68 \\
UEAF & 44.78$\pm$0.31 & 38.75$\pm$0.52 & 33.21$\pm$0.58 & 28.91$\pm$0.60 & 24.55$\pm$0.44\\
AGC\_IMC & 39.46$\pm$2.12 & 33.08$\pm$2.37 & 27.88$\pm$1.28 & 25.35$\pm$1.34 & 21.95$\pm$0.48\\
SAGF\_IMC & 25.89$\pm$1.69 & 25.59$\pm$1.01 & 22.89$\pm$1.29 & 21.14$\pm$0.83 & 20.07$\pm$0.54\\
LRTL & 33.77$\pm$1.42 & 29.04$\pm$1.51 & 25.34$\pm$1.26 & 21.75$\pm$0.57 & 21.18$\pm$0.77\\
PIMVC & 43.70$\pm$0.45 & 39.29$\pm$0.76 & 35.47$\pm$0.70 & 30.92$\pm$1.19 & 24.54$\pm$0.72\\
CRG\_IMSC & 48.83$\pm$1.01 & \textbf{43.19}$\pm$1.36 & \textbf{37.91}$\pm$0.56 & \textbf{30.96}$\pm$0.79 & \textbf{24.97}$\pm$0.78 \\
\hline
\end{tabular}
\label{tab:wiki3}
\end{table*}

\begin{table*}[!h]
\renewcommand{\arraystretch}{1.3}
\centering
\caption{Precision(\%)(mean and standard deviation) under different missing rates on Wikipedia dataset}
\footnotesize
\begin{tabular}{cccccc}
\hline
\bfseries Method\verb|\|MR & 10\% & 30\% & 50\% & 70\% & 90\% \\
\hline
BSV & 47.39$\pm$0.43 & 32.61$\pm$1.15 & 24.11$\pm$0.38 & 18.32$\pm$0.33 & 14.68$\pm$0.21 \\
MultiNMF & 35.09$\pm$1.13 & 31.25$\pm$0.65 & 26.89$\pm$0.91 & 24.01$\pm$0.63 & 21.14$\pm$0.99 \\
Spec-Pair & 39.07$\pm$1.47 & 34.89$\pm$0.94 & 30.95$\pm$1.36 & 25.32$\pm$1.96 & 19.29$\pm$3.15 \\
Spec-Cent & 42.57$\pm$0.51 & 36.95$\pm$0.75 & 31.96$\pm$0.71 & 27.13$\pm$0.98 & 22.19$\pm$0.79 \\
PVC & 46.23$\pm$4.37 & 36.42$\pm$5.16 & 30.13$\pm$1.40 & 25.30$\pm$1.98 & 20.47$\pm$0.96 \\
DAIMC & 44.44$\pm$0.99 & 31.60$\pm$0.71 & 17.80$\pm$0.98 & 12.10$\pm$0.33 & 11.13$\pm$0.14 \\
SRLC & 37.28$\pm$1.57 & 32.83$\pm$0.75 & 28.55$\pm$0.82 & 24.41$\pm$0.97 & \textbf{21.05}$\pm$0.71 \\
UEAF & 44.98$\pm$0.31 & \textbf{38.62}$\pm$0.76 & 32.87$\pm$0.71 & 28.39$\pm$0.66 & 23.47$\pm$0.66\\
AGC\_IMC & 35.70$\pm$2.64 & 28.58$\pm$3.30 & 22.72$\pm$1.87 & 20.81$\pm$2.28 & 17.63$\pm$1.14\\
SAGF\_IMC & 19.60$\pm$1.46 & 19.62$\pm$1.73 & 17.75$\pm$1.73 & 16.02$\pm$1.33 & 15.65$\pm$0.58\\
LRTL & 34.27$\pm$1.38 & 29.44$\pm$1.71 & 25.62$\pm$1.01 & 21.86$\pm$0.51 & 20.97$\pm$0.76\\
PIMVC & 43.69$\pm$0.50 & 39.15$\pm$0.81 & 35.26$\pm$0.63 & 30.46$\pm$1.24 & 24.26$\pm$0.89\\
CRG\_IMSC & \textbf{48.02}$\pm$0.53 & 36.86$\pm$1.91 & \textbf{35.30}$\pm$0.73 & \textbf{28.45}$\pm$1.19 & 20.65$\pm$1.31 \\
\hline
\end{tabular}
\label{tab:wiki4}
\end{table*}

\subsection{Compared Methods}
CRG\_IMSC is compared with 10 clustering methods, i.e., MultiNMF \cite{ref10}, Spec-Pair \cite{ref34}, Spec-Cent \cite{ref34}, PVC \cite{ref24}, DAIMC \cite{DAIMC}, SRLC \cite{ref25}, UEAF \cite{ref22}, AGC\_IMC \cite{ref39}, SAGF\_IMC \cite{ref27}, LRTL \cite{Chen2023}, PIMVC \cite{Deng2023} and Best Single View (BSV). For BSV, we perform $K$-means in each view and employ the best result. For CRG\_IMSC and SRLC, the clustering results are generated by the algorithms directly. For the other 8 methods, $K$-means is performed on the features extracted by the corresponding algorithms. With different initializations, $K$-means will generate different clustering results. Thus, in the experiments, it is performed 20 times and the average value of NMI, Accuracy, F-score and Precision are displayed.

\subsection{Experimental Results and Analysis}
Considering BSV, MultiNMF, Spec-Pair and Spec-Cent are not incomplete multi-view clustering methods, for these four methods, we use the average value to fill the missing views. The parameters of all the compared methods are set as the original papers. For CRG\_IMSC, the hyper-parameters $\alpha$ and $\beta$ are selected from $\{1,10,100\}$ and $\{0.01,0.1,1\}$, respectively. The parameter setting of CRG\_IMSC will be analyzed in the following section.

On Wikipedia dataset, in order to evaluate the clustering performance of CRG\_IMSC for incomplete multi-view data, we let 10\% to 90\% samples have missing views, with the interval of 20\%. These incomplete samples are randomly selected, and at least one view of the samples is existing. We conduct the random selection 10 times and adopt the average clustering results. Table~\ref{tab:wiki1}, Table~\ref{tab:wiki2}, Table~\ref{tab:wiki3} and Table~\ref{tab:wiki4} list mean and standard deviation of NMI, Accuracy, F-score and Precision under different missing rates, respectively. From these tables, we can see that, CRG\_IMSC almost has the best or the second best performances under all missing rates. CRG\_IMSC obtains the highest Accuracy and F-score, except for the F-score under missing rate 10\%. Although PVC has the highest F-score under missing rate 10\%, its standard deviation is also the biggest, which indicates that the result is not stable. For NMI, BSV obtains the best performance. However, NMI of CRG\_IMSC is just slightly lower than BSV.

\begin{figure*}[!h]
\centering
\subfloat[3Sources]{\includegraphics[width=7cm]{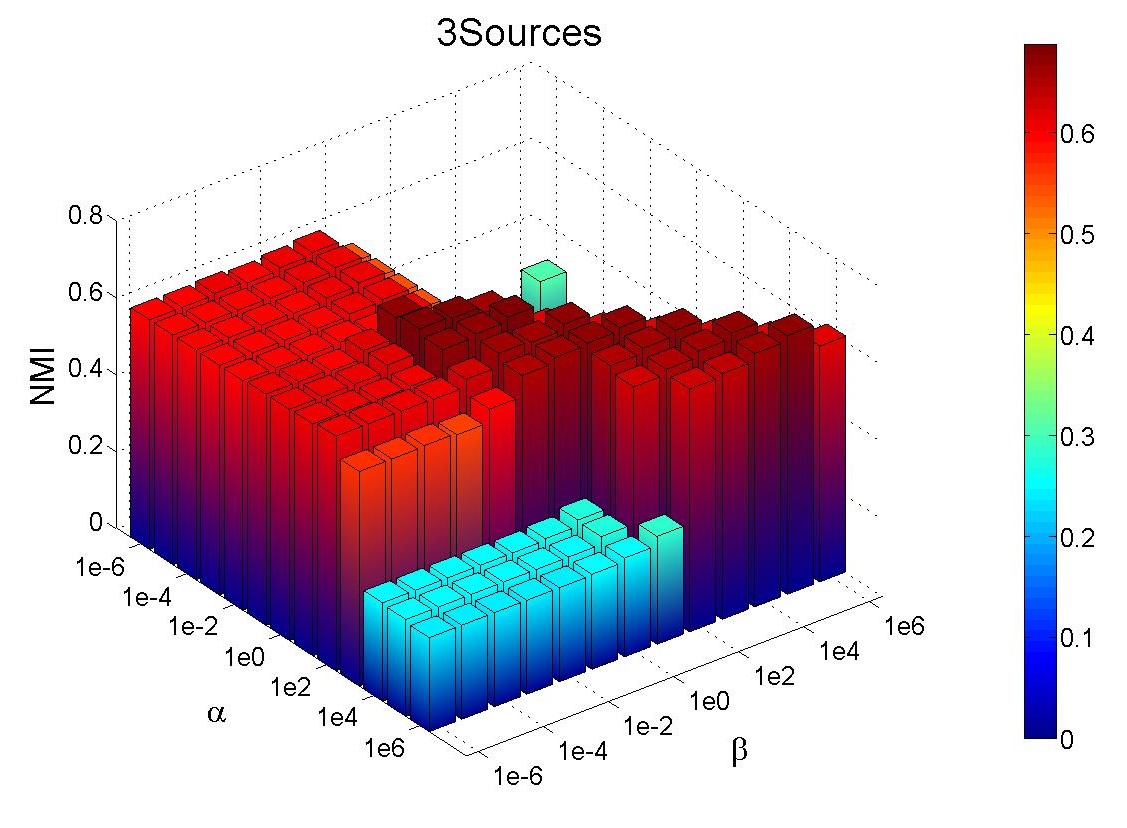}}
\hfil
\subfloat[BBCSport]{\includegraphics[width=7cm]{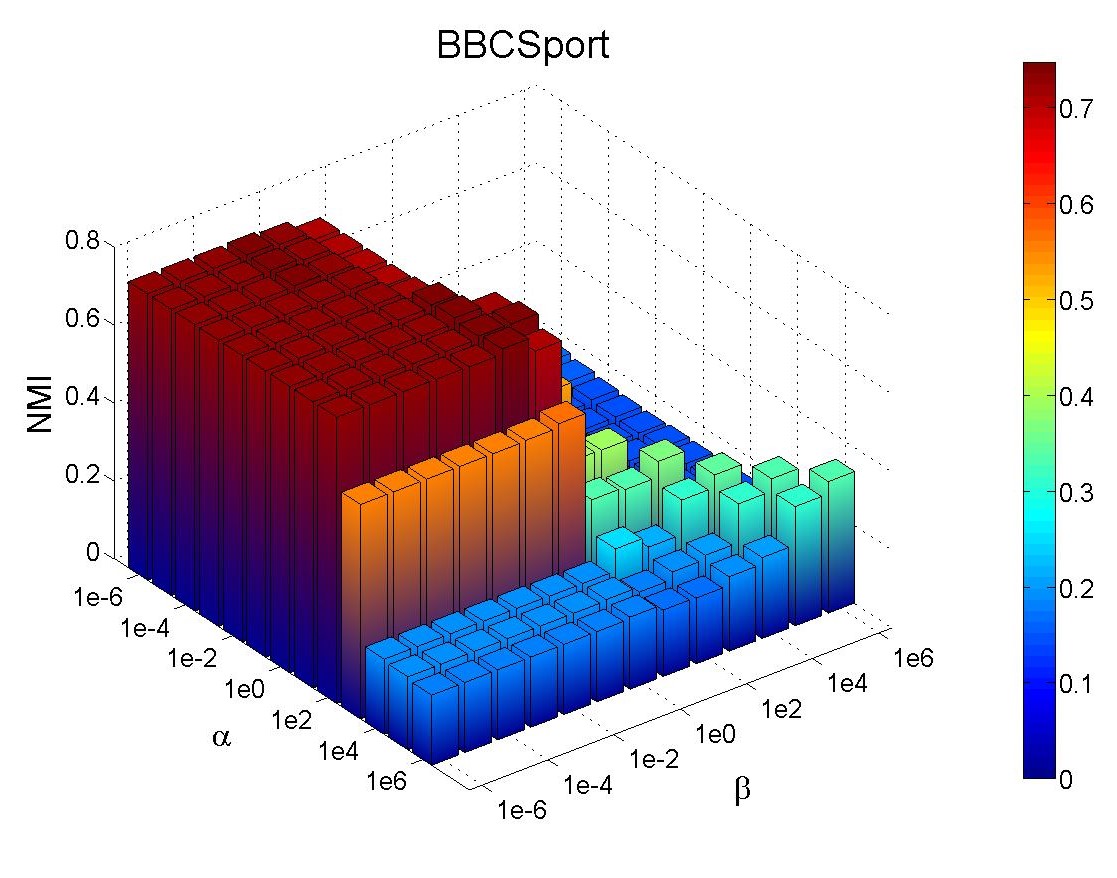}}
\caption{NMI of CRG\_IMSC versus parameters $\alpha$ and $\beta$ on 3Sources and BBCSport datasets}
\label{fig:para}
\end{figure*}

\begin{table}
\renewcommand{\arraystretch}{1.3}
\centering
\caption{NMI, Accuracy, F-score and Precision(\%) on 3Sources dataset}
\footnotesize
\begin{tabular}{ccccc}
\hline
\bfseries Method\verb|\|Index & NMI & Accuracy & F-score & Precision\\
\hline
 BSV & 46.14 & 56.67 & 48.45 & 43.81\\
 MultiNMF & 43.85 & 51.18 & 36.59 & 27.47\\
 Spec-Pair & 44.17 & 57.66 & 42.57 & 40.38\\
 Spec-Cent & 51.81 & 63.20 & 49.60 & 48.34\\
 DAIMC & 39.66 & 52.96 & 41.97 & 38.57\\
 SRLC & 51.14 & 70.67 & 58.37 & 57.00\\
 UEAF & 62.51 & 65.95 & 58.33 & 55.91\\
 AGC\_IMC & 70.16 & 79.19 & 70.96 & 61.96\\
 SAGF\_IMC & 64.38 & 73.13 & 64.98 & 55.82\\
 LRTL & 51.35 & 40.31 & 34.50 & 40.89 \\
 PIMVC & 81.67 & 72.54 & 68.22 & 70.20 \\
 CRG\_IMSC & \textbf{70.87} & \textbf{85.10} & \textbf{73.23} & \textbf{74.92}\\
\hline
\end{tabular}
\label{tab:3Sources}
\end{table}

\begin{table}
\renewcommand{\arraystretch}{1.3}
\centering
\caption{NMI, Accuracy, F-score and Precision(\%) on BBC dataset}
\footnotesize
\begin{tabular}{ccccc}
\hline
\bfseries Method\verb|\|Index & NMI & Accuracy & F-score & Precision\\
\hline
 BSV & 29.22 & 47.25 & 38.98 & 33.34\\
 MultiNMF & 40.82 & 55.13 & 43.65 & 36.37\\
 Spec-Pair & 1.19 & 25.64 & 25.06 & 20.72\\
 Spec-Cent & 13.60 & 35.01 & 33.83 & 22.64\\
 DAIMC & 35.49 & 49.79 & 43.32 & 38.95\\
 SRLC & 51.15 & 72.81 & 60.56 & 60.10\\
 UEAF & 63.03 & 78.78 & 67.43 & 63.72\\
 AGC\_IMC & 75.30 & 86.24 & 81.24 & 78.21\\
 SAGF\_IMC & 71.86 & 78.38 & 75.70 & 70.99\\
 LRTL & 61.39 & 46.69 & 41.82 & 45.73 \\
 PIMVC & 87.01 & 82.87 & 77.06 & 81.07 \\
 CRG\_IMSC & \textbf{75.39} & \textbf{90.47} & \textbf{83.13} & \textbf{82.07}\\
\hline
\end{tabular}
\label{tab:BBC}
\end{table}

\begin{table}
\renewcommand{\arraystretch}{1.3}
\centering
\caption{NMI, Accuracy, F-score and Precision(\%) on BBCSport dataset}
\footnotesize
\begin{tabular}{ccccc}
\hline
\bfseries Method\verb|\|Index & NMI & Accuracy & F-score & Precision\\
\hline
BSV & 18.85 & 39.97 & 35.79 & 30.07\\
MultiNMF & 43.24 & 53.49 & 47.78 & 43.52\\
Spec-Pair & 6.67 & 29.28 & 29.53 & 23.79\\
Spec-Cent & 6.23 & 29.53 & 29.74 & 23.64\\
DAIMC & 24.36 & 40.68 & 36.80 & 31.76\\
SRLC & 53.26 & 68.66 & 63.60 & 60.71\\
UEAF & 57.30 & 74.61 & 63.55 & 67.08\\
AGC\_IMC & 72.97 & 82.75 & 80.01 & 77.06\\
SAGF\_IMC & 66.09 & 70.62 & 65.49 & 52.52\\
LRTL & 43.13 & 37.56 & 25.44 & 39.04 \\
PIMVC & 84.29 & 79.39 & 76.55 & 77.12 \\
CRG\_IMSC & \textbf{76.91} & \textbf{90.64} & \textbf{81.95} & \textbf{83.67}\\
\hline
\end{tabular}
\label{tab:BBCSport}
\end{table}

3Sources, BBC and BBCSport have missing view originally. 3Sources has 3 views, and BBC and BBCSport both have 4 views. Because PVC can only handle two-view data, we did not perform it on these three datasets. Table~\ref{tab:3Sources}, Table~\ref{tab:BBC} and Table~\ref{tab:BBCSport} list four clustering validity indexes on 3Sources, BBC and BBCSport datasets, respectively. It can be found that, on three datasets, CRG\_IMSC all obtains the best performances, no matter which clustering validity index is adopted. Incomplete multi-view spectral clustering method AGC\_IMC also performs well on three datasets. Nevertheless, for Accuracy and Precision, our CRG\_IMSC is significantly better than AGC\_IMC.

 \begin{figure*}[!h]
\centering
\subfloat[BBC]{\includegraphics[height=5cm,width=8cm]{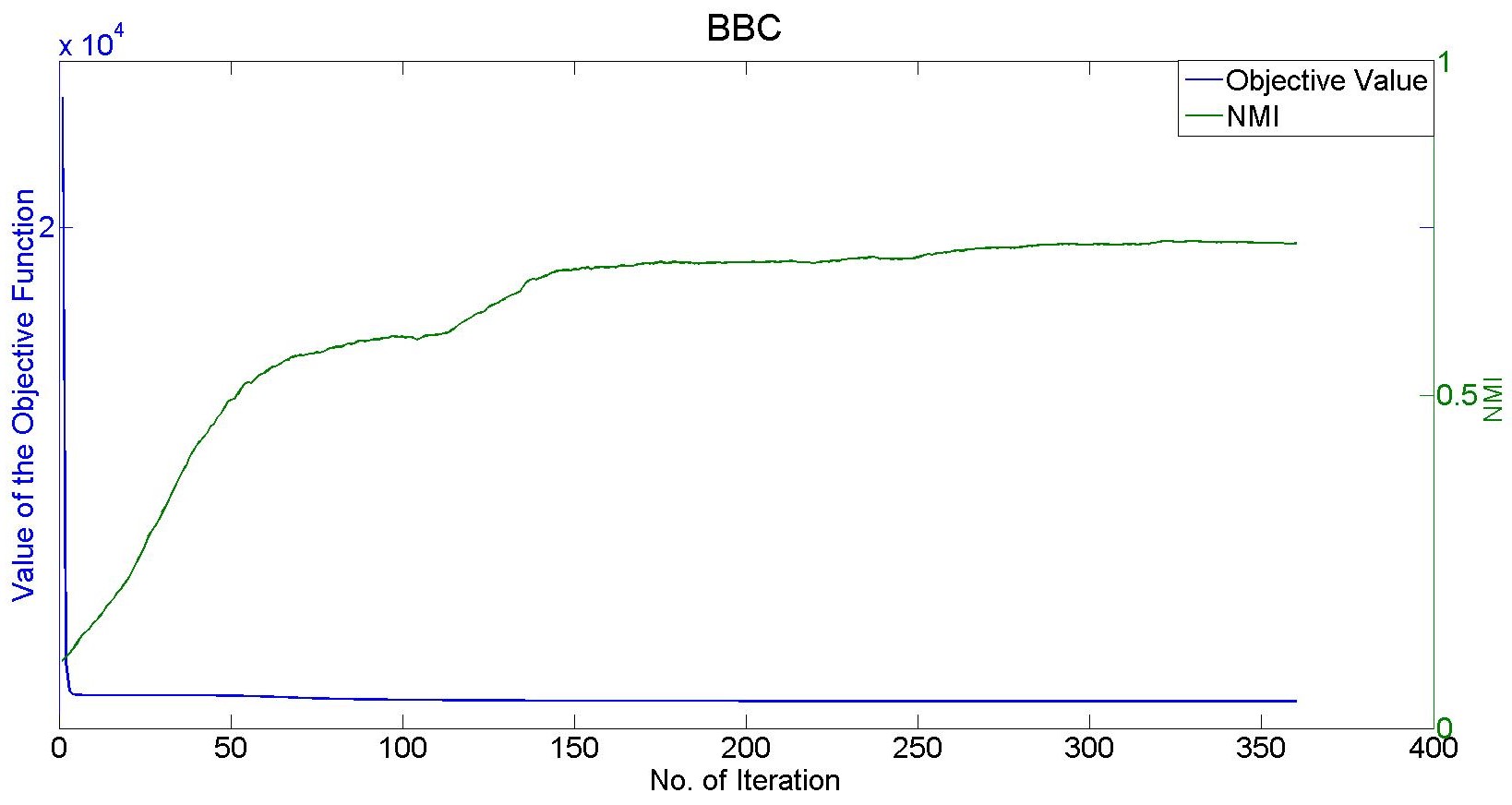}}
\hfil
\subfloat[BBCSport]{\includegraphics[height=5cm,width=8cm]{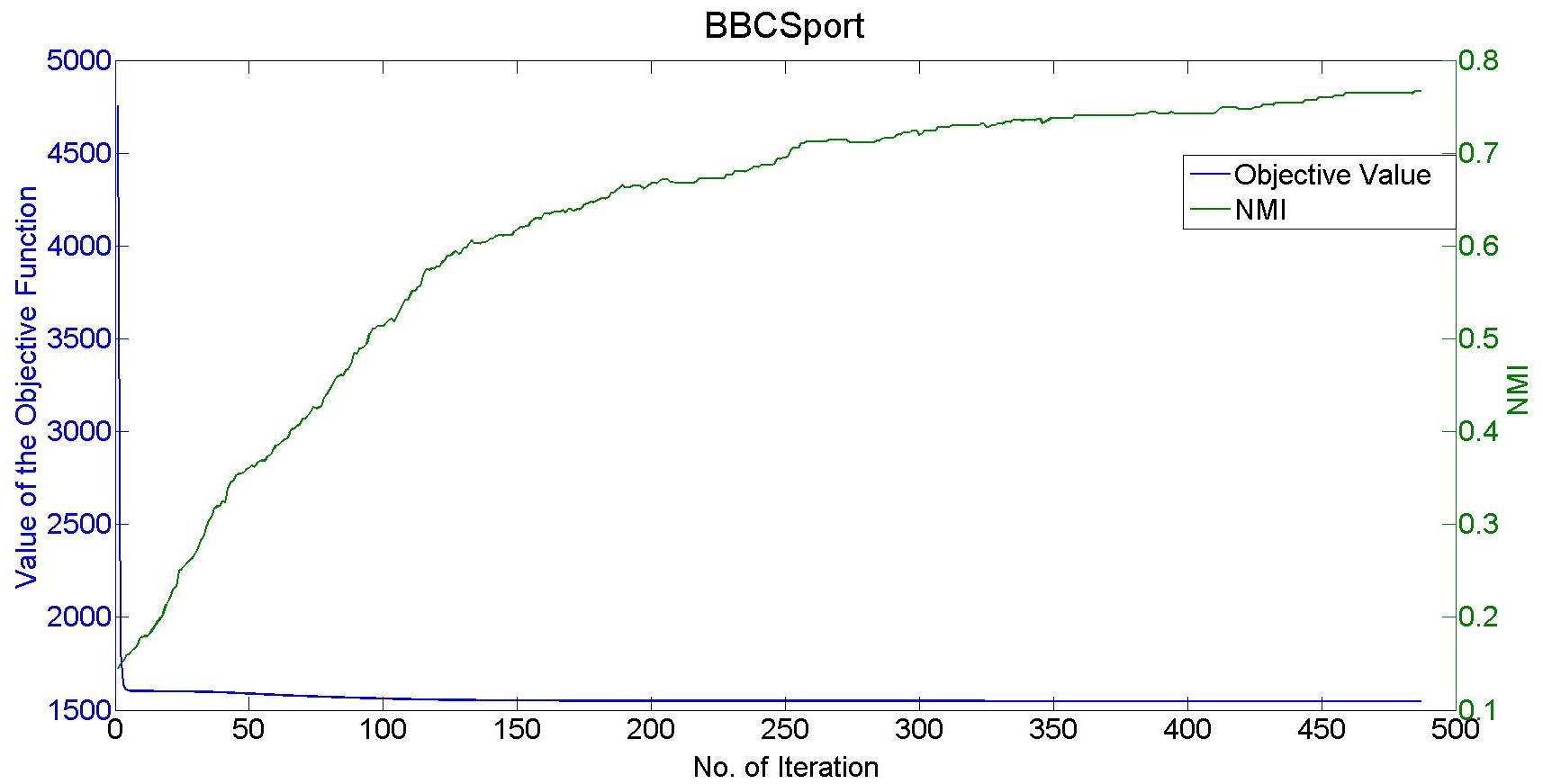}}
\caption{The value of objective function of CRG\_IMSC and NMI in each iteration}
\label{fig:conver}
\end{figure*}

\subsection{Parameter Study}
For two parameters $\alpha$ and $\beta$ of CRG\_IMSC, to find the optimal values of them, we let them vary in the range of $\{10^{-6},10^{-5},10^{-4},\dots,10^5,10^6\}$ and calculate NMI under different parameter settings. Experiments are performed on 3Sources and BBCSport datasets. Fig.~\ref{fig:para} shows NMI with different combinations of $\alpha$ and $\beta$. From Fig.~\ref{fig:para}, we can find that, on both datasets, CRG\_IMSC obtains the highest NMI with $\alpha\in\{1,10,100\}$ and $\beta\in\{0.01,0.1,1\}$. It can be also seen from Fig.~\ref{fig:para} that, the values of $\alpha$ and $\beta$ must be matching. With the combination of big $\alpha$ and small $\beta$, or the combination of small $\alpha$ and big $\beta$, CRG\_IMSC both obtains terrible performances. These two hyper-parameters balance Laplacian term and consistency constraint term. Either term is too big, the clustering performance will decrease significantly.

\subsection{Convergence Study}
The convergence of CRG\_IMSC algorithm is evaluated on BBC and BBCSport datasets. In each iteration, the value of objective function is calculated, and NMI is also recorded. Fig.~\ref{fig:conver} shows the results, where the blue curves indicate the value of objective function and the green curves indicate NMI. From Fig.~\ref{fig:conver}, we can see that, along with the increasement of iteration, the value of objective function decreases, while NMI increases. In the beginning, the value of objective function decreases quickly and then it becomes stable. After hundreds of iterations, the algorithm converges, and almost at the same time the maximum NMI is obtained.

\section{Conclusion}
This paper proposes a new incomplete multi-view spectral clustering method CRG\_IMSC. Unlike the traditional spectral clustering method, CRG\_IMSC does not need to perform $K$-means algorithm after feature extraction, and the clustering result can be generated from the feature representation directly. For CRG\_IMSC, the clustering result is used to construct the connectivity matrix, and it is employed in self-representation to guide multi-view spectral clustering. A novel iterative algorithm is exploited to solve the optimization problem of CRG\_IMSC, and it is proved to be convergent in theory. CRG\_IMSC is compared with state-of-the-art multi-view clustering method on four public benchmarks. Experimental results demonstrate the superior effectiveness of CRG\_IMSC. The optimal ranges of two parameters of CRG-IMSC are also given by parameter study experiments. Besides, the convergence of CRG\_IMSC algorithm is evaluated in the experiments.

In the future, we will generalize our method to large scale multi-view data. For large scale data, the efficiency of the algorithm is critical. Thus, we will try to develop more efficient algorithm to solve the optimization problem. In addition, now  hyper-parameters are set empirically in our method, more precise parameter setting method will also be explored in the future work.

\section*{Acknowledgments}
This work is supported by National Natural Science Foundation of China (62076096, 62476166), Shanghai Pujiang Program (22PJD029) and Shanghai Municipal Project (2051110090).




%

\bibliographystyle{IEEEtran}
\bibliography{reference}







\begin{IEEEbiography}[{\includegraphics[width=1in,height=1.25in,clip,keepaspectratio]{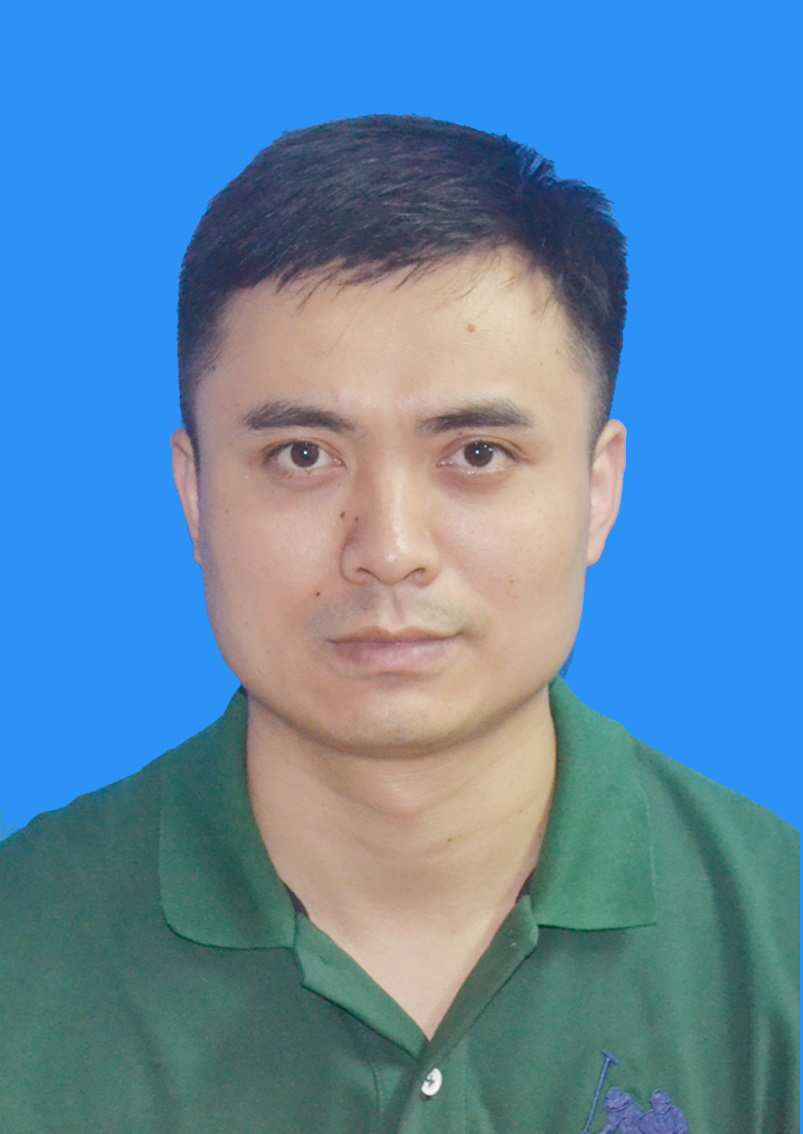}}]{Jun Yin}
received the B.S. degree in Mathematics and the Ph.D. degree in Pattern Recognition and Intelligent System from Nanjing University of Science and Technology, Nanjing, China in 2006 and 2011 respectively.

He is an associate professor at the College of Information Engineering, Shanghai Maritime University. From 2016 to 2020, he was a postdoctoral fellow at the School of Computer Science and Technology, East China Normal University. From 2019 to 2020, he was a visiting scholar with the Department of Electrical and Computer Engineering, University of Pittsburgh, Pittsburgh, USA. His research interests include multi-view learning, clustering, dimension reduction, manifold learning, etc. He has published more than 50 papers at peer-reviewed journals and conferences, such as IEEE TRANSACTIONS ON KNOWLEDGE AND DATA ENGINEERING, IEEE TRANSACTIONS ON NEURAL NETWORKS AND LEARNING SYSTEMS, PATTERN RECOGNITION, INFORMATION FUSION, AAAI and ACMMM.

Dr. Yin is on the Editorial Board of Expert Systems With Applications, and Neural Processing Letters.
\end{IEEEbiography}

\begin{IEEEbiography}[{\includegraphics[width=1in,height=1.25in,clip,keepaspectratio]{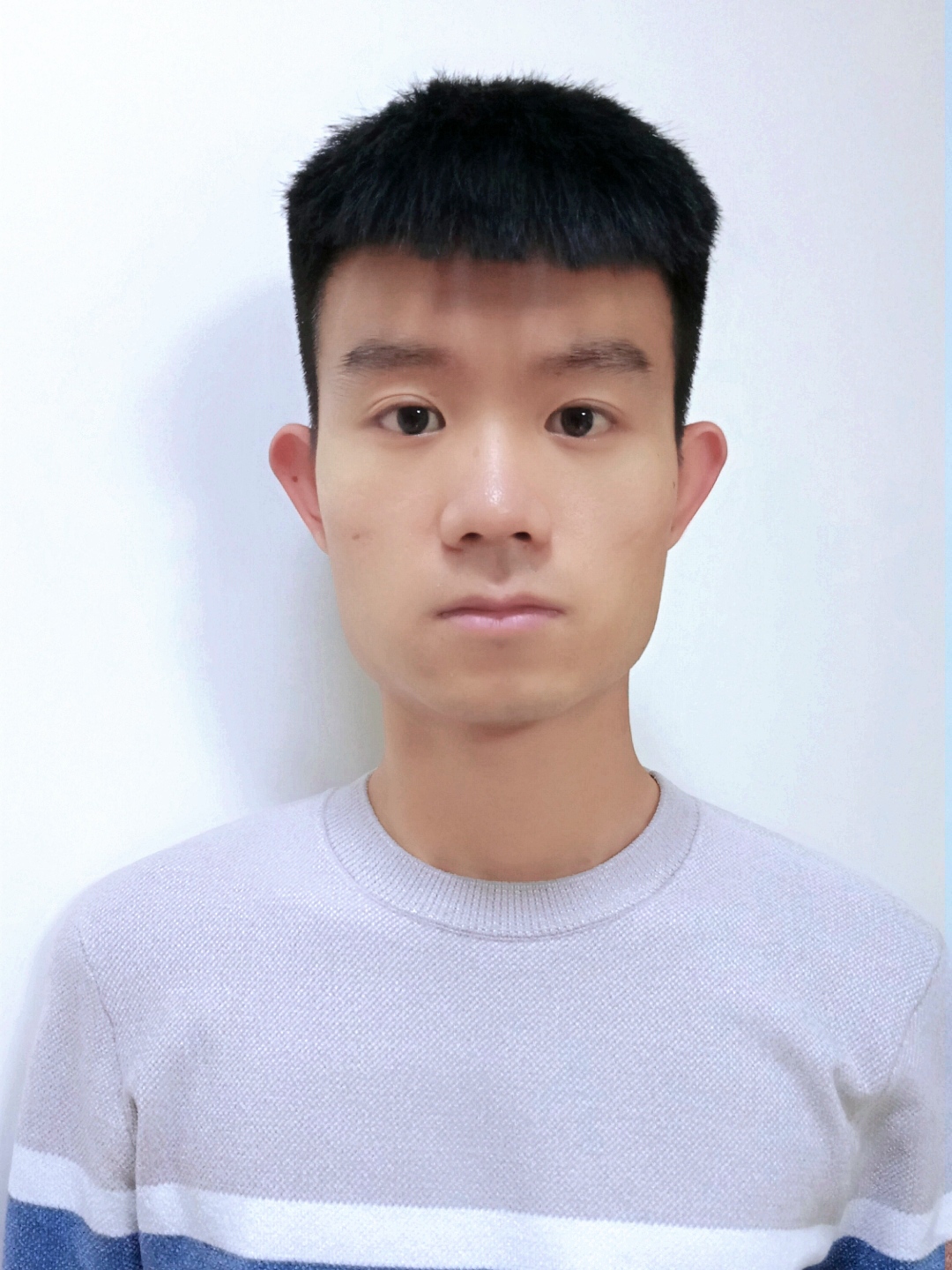}}]{Runcheng Cai}
received the B.S. degree in Electronic Information Science and Technology from Dezhou University, Shandong, China in 2019, and the M.S. degree in Computer Science and Technology at the College of Information Engineering, Shanghai Maritime University, Shanghai, China in 2023. His current research interest is multi-view clustering.
\end{IEEEbiography}

\begin{IEEEbiography}[{\includegraphics[width=1in,height=1.25in,clip,keepaspectratio]{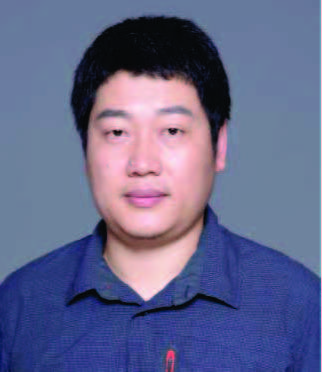}}]{Shiliang Sun} (Senior Member, IEEE) received the B.E. degree in automation from Beijing University of Aeronautics and Astronautics, Beijing, China, in 2002, and the Ph.D. degree in pattern recognition and intelligent systems from Tsinghua University, Beijing, in 2007.

He is currently a Professor with the Department of Automation, Shanghai Jiao Tong University, and the Founding Head of the Pattern Recognition and Machine Learning Research Group, East China Normal University, Shanghai, China. In 2004, he was named a Microsoft Fellow. From 2009 to 2010, he was a Visiting Researcher with the Department of Computer Science, Centre for Computational Statistics and Machine Learning, University College London, London, U.K. In 2014, he was a Visiting Researcher with the Department of Electrical Engineering, Columbia University, New York, NY, USA. From 2011 to 2023, he was a Full Professor with East China Normal University. His current research interests include causal inference, reinforcement learning, reasoning theory and technology, and multiview learning. His research results have expounded in more than 100 publications at peer-reviewed journals and conferences, such as JMLR, IEEE TRANSACTIONS ON PATTERN ANALYSIS AND MACHINE INTELLIGENCE, IEEE TRANSACTIONS ON NEURAL NETWORKS AND LEARNING SYSTEMS, IEEE TRANSACTIONS ON CYBERNETICS, ICML, and NeurIPS, and multiple academic monographs and textbooks, such as Multiview Machine Learning and Pattern Recognition and Machine Learning.

Prof. Sun has served or is currently an Editor of multiple international journals, including IEEE TRANSACTIONS ON PATTERN ANALYSIS AND MACHINE INTELLIGENCE and IEEE TRANSACTIONS ON NEURAL NETWORKS AND LEARNING SYSTEMS.
\end{IEEEbiography}

\vfill

\end{document}